# MUTUAL TRANSFORMATION OF INFORMATION AND KNOWLEDGE

Olegs Verhodubs

oleg.verhodub@inbox.lv

*Abstract*-**Information and knowledge are transformable into each other. Information transformation into knowledge by the example of rule generation from OWL (Web Ontology Language) ontology has been shown during the development of the SWES (Semantic Web Expert System). The SWES is expected as an expert system for searching OWL ontologies from the Web, generating rules from the found ontologies and supplementing the SWES knowledge base with these rules. The purpose of this paper is to show knowledge transformation into information by the example of ontology generation from rules.**

## I. INTRODUCTION

The development of the Web during the last few decades has led to the accumulation of large amount of information in the common information environment. In 1997 search engines claimed to index from 2 million to 100 million web documents [1]. The number of documents, indexed by web search engines, is increasing from year to year. For example, Google has an index of over 30 trillion web pages now [2]. The information, indexed by web search engines, is not homogeneous, and it is presented in the Web in different forms. Web pages, different documents, pictures, ontologies, archives and others are these forms of information in the Web. In general, it is necessary to distinguish information, data and knowledge. This difference of data, information and knowledge will be explained hereinafter, but here it can be argued that this difference is a serious obstacle to the full use of the potential of the Web. There are several ways to eliminate this obstacle. For instance, it is theoretically possible to develop one unified language for the Web in order to represent data, information or knowledge, but in practice it is hardly feasible in terms of its use. In this connection, the way of mutual transformations of data, information and knowledge is the most suitable in terms of implementation. This work will not start from the scratch, because some types of these transformations have already been developed. Generation of rules from OWL (Web Ontology Language) ontology was investigated as part of SWES (Semantic Web Expert System) development [3], [4]. The development of the SWES is the main purpose of the research. An expert system, which is based on the Semantic Web technologies, is meant under the SWES. The SWES is being developed as the system, which looks for OWL ontologies from the Web, generates rules from the OWL ontologies and supplements its own knowledge base with these rules. These actions, as well as communication with the user, will give an opportunity to get the exact answer to the user's request. This style of work is significantly different from the existing systems in the Web, which as a result provide a list of links to resources that may contain the answer. Moreover, it is expected that the SWES will extract more knowledge from the Web than existing systems are able to do so.

It is logical to assume that the task of ontology generation from rules, which is the opposite task to the task of rule generation from ontology, is realizable. The main reason for confidence in the fact that this so, is the essence of data, information and knowledge. Point is that data, information and knowledge are a single entity having different forms. Different forms of a single entity are perceived as entities with fundamental differences. Apparently this is due to limitations in the human perception of reality. This paper will fit this gap. The purpose of this paper is to understand what are data, information and knowledge, as well as to find out their differences, and in addition to present the way of ontology generation from rules.

This paper structured as follows. The next section clarifies terms such as data, information and knowledge. The third section introduces the new conception of data, information and knowledge. The fourth section describes the way of ontology generation from rules. Finally, the conclusions follow.

## II. BACKGROUND

It is necessary to clarify the definitions of such concepts as data, information and knowledge. This is especially important, because many people, including professionals, are confused in these terms and use them interchangeable [5].

There are different definitions of data, information and knowledge. According to Ackoff data are symbols that represent properties of objects, events and their environment, and they are the results of observation [7]. Information is in descriptions and answers to questions that begin with the words who, what, when, how many; information is inferred from data [7]. Knowledge is something that makes possible to transform information to instructions and is obtained by extracting from experience [7]. There are alternative definitions of data, information and knowledge, too. According to [8] data concern with observation and raw facts. They are useless without an additional processing namely comparing, inferring, filtering etc [8]. In turn, the processed data is known as information [8]. Knowledge can be classified in many different ways: tacit knowledge, explicit knowledge, factual knowledge, procedural knowledge, commonsense knowledge, domain knowledge, meta knowledge and so on [8]. Domain knowledge is valid knowledge for a specified domain. Specialists and experts develop their own domain knowledge and use it for problem solving [8]. Meta knowledge can be defined as knowledge about knowledge [8]. Common sense knowledge is a general purpose knowledge expected to be present in every normal human being. Common sense ideas tend to relate to events within human experience [8]. Heuristic knowledge is a specific rule-of-thumb or argument derived from experience [8]. Explicit knowledge is knowledge that can be easily expressed in words/numbers and shared in the form of data, scientific formulae, product specifications, manuals, and universal principles. It is more formal and systematic [8]. Tacit knowledge is the knowledge stored in subconscious mind of experts and not easy to document. It is highly personal and hard to formalize, and hence difficult to represent formally in system. Subjective insights, intuitions, emotions, mental models, values and actions are examples of tacit knowledge [8]. Many other definitions identify data as representation of facts or ideas in a formalized manner [9], as representation of facts about the world [9], or the representation of concepts and other entities [9]. Other definitions of information formulate it as a message that contains relevant meaning, implication, or input for decision and/or action [9], and specify that information is given meaning by way of context [9]. Knowledge in other definitions is identified as the the cognition or recognition (know-what), capacity to act (know-how), and understanding (know-why) that resides or is contained within the mind or in the brain [9]. All these alternative definitions and a lot of others reffered to data, information and knowledge are stacked in a DIKW (Data, Information, Knowledge, Wisdom) hierarchy or pyramid, proposed by Ackoff [7]. This hierarchy means transformation process from the lowest level, which is data, to the highest level, which is knowledge. Ackoff implies the only one direction of transformation namely from the lowest level to the highest level.

### III. NEW CONCEPTION OF DATA, INFORMATION AND KNOWLEDGE

The Ackoff's understanding of data, information and knowledge, stacked in a DIKW hierarchy, where there is the only one direction of transformation namely from the lowest level to the highest level has a lot of criticism [10]. Indeed, data is the primary element of the DIKW hierarchy, and other elements of this hierarchy as information and knowledge are derived from the data that is information is derived from data, whereas knowledge is derived from information. But a human cognition cannot see simple facts without these facts being part of its current meaning structure, and this means there are no isolated pieces of simple facts unless someone has created them using his or her knowledge [11]. Thus, data can emerge only if a meaning structure is first fixed and then used to represent information [11]. Consequently, it is possible to conclude that the DIKW hierarchy is not adequate, or it is not quite adequate. One of the existing proposals to make the hierarchy of data-information-knowledge that is DIKW hierarchy more adequate is to turn out this hierarchy other way around in such a way that data emerges last – only after there is knowledge and information available [11]. Following this, instead of being raw material for information, data emerges as a result of adding value to information by putting it into a form that can be automatically processed [11]. One of the illustrative examples, where data emerges last, after a meaning structure is fixed, is a semantically well defined computer database, where information is stored in.

In truth, the proposal of the turned out DIKW hierarchy is not exhaustive in the sense that this proposal would cancel the standard DIKW hierarchy. The standard DIKW hierarchy and its turned variant are both acceptable and are true. The difference between the standard DIKW hierarchy and its turned variant is in the starting point whether this is data or knowledge. Considering the relationship between the categories of data, information and knowling as well as admissibility of multidirectional transformations from data to knowledge and in the opposite direction that is from knowledge to data, it is possible to conclude that data, information and knowledge in total represent something in common that is a single substance, which may be perceived and interpreted differently namely as data, information and knowledge depending on the purpose of the perception or interpretation. According to Boley and others facts are derivation rules without premises [12]. Hence data are also derivation rules without premises, because data akin to facts [10]. It is possible to assume that any data is rule at least in one case. For example, the statement or information "I do not believe in actions without goals." can be perceived as data and knowledge (rules). Sensing this statement as data, it is possible to determine that there is someone, someone has some belief, and this belief is that there is no actions without goals. This statement can be perceived as at least one rule namely if there are actions, then there are goals of these actions. If this rule were not true, it would be true for someone, who expressed its belief. In this sense, this rule would be true in one case.

Following the logic of different perception depending of its purpose, it is possible to state the presence of the purpose is a major limiting factor of human perception. It is necessary to notice that information is the closest name of the single substance, which combines data, knowledge and information itself. Any information contains data and knowledge that can be extracted by means of ordinary analisys if it is needed. It is only necessary to know that a set of signals, also named as information, has its internal order or system. In other words information has to be true.

IV. ONTOLOGY GENERATION FROM RULES

Presence of true information gives an opportunity to get true data and knowledge (rules). At the same time, it is logical to assume that if there is consistent knowledge (rules), then these rules can be used to generate consistent data, information and knowledge in another form of its representation, such as ontology. Ontology is an explicit and formal specification of a conceptualisation of a domain of interest [13]. Ontologies consist of concepts or classes, properties, instances and axioms [14]. OWL (Web Ontology Language) is a typical example of ontology language [15]. The OWL language provides mechanisms for creating all the components of an ontology: concepts, instances, properties (or relations) and axioms. Two sorts of properties can be defined: object properties and datatype properties [14]. Object properties relate instances to instances, and datatype properties relate instances to datatype values, for example, text strings or numbers. Concepts can have super and subconcepts, thus providing a mechanism for subsumption reasoning and inheritance of properties. Finally, axioms are used to provide information about classes and properties, for example, to specify the equivalence of two classes or range of a property [14].

It is necessary to correct the understanding of the ontology, but rather about the components of the ontology in order to bring this understanding to real life. This is necessary, because the understanding of the ontology as concepts, instances, axioms and two types of properties is closer to the programmer's understanding, but not to real life or in other words to human understanding. It is much more productive to split ontology into categories of concepts, properties, relations and instances, where concepts and instances are identical to the previous categorizations, whereas properties are attributes of concepts and relations are links between concepts or instances. Axioms in the old categorization are removed in the new categorization and instead standard types of links are added. If Object property is an arbitrary link as, for example, "hasParent", "canFly" and so on, then standard types of links include such links as "equivalentClass", "subClassOf" and "complementOf". There are other axioms, which can be added to the new categorization as standard link types, but this is the nex task, but now it is necessary to show the principle of new categorization. Let us illustrate old and new categorizations in Table I.

TABLE I. Categorization of ontology components.

| Nr. p.k. | Component name | Categorization | |
|---|---|---|---|
| | | Old | New |
| 1 | Concept | class | class |
| 2 | Instance | class exemplar | class exemplar |
| 3 | Property | Datatype property | Datatype property |
| | | Object property | |
| 4 | Relation | - | Object property |
| | | | equivalentClass |
| | | | subClassOf |
| | | | complementOf |
| 5 | Axiom | Component info | - |

New categorization of ontology components is more natural and clearer, because it allows to imagine and picture any ontology by means of grafic elements as graph is pictured. Such notation had already been used, but not described, when the task of rule generation from ontology had been discussed [3]. Now it is possible to examine each rule from the perspective of ontology components. Boley and others consider the facts as derivation rules without premises [12]. I think that this definition of facts can be precized and expanded at the same time. First, facts are rules without premises, but not only derivation rules without premises. Second, facts are not only rules without premises that is conclusions. Facts are rules without conclusions that is premises. In other words, premises and conclusions are facts. Of course, premises and conclusions are facts, which belong to different categories, but outside the rule premise and conclusion are something that can exist and exists, and this something is fact. Here is a place for further research, which is possible will change our understanding about what is rule and others, however this is another research. So, rule premises and conclusions are facts, but they are facts in the sense that they exist, although these facts may be not facts in the sense of instances, but facts in the sense of concepts. For example, the rule:

**IF** Kitty **THEN** Cat,

contains two facts: Kitty, Cat. The fact "Kitty" means the instance that is the name of some cat, but the fact "Cat" means the concept.

Simplistically the process of ontology generation from rules begins with extracting of facts and then follows sorting of extracted facts by categories of ontology components. In general, this is the inverse process of the rule generation from ontology task [16], however it has its obstacles. The main obstacle is that ontology generation from rules is a creative process to some extent in the sense that one rule can be transformed to different ontology code fragments. For example, the following rule:

**IF** Wheel **and** Engine **THEN** Car,

can be transformed to different OWL ontology code fragments. The first possible code fragment is:

```
<owl:Class rdf:ID="#Car"/>
 <owl:DatatypeProperty rdf:ID="Wheel">
  <rdfs:domain rdf:resource="#Car"/>
  <rdfs:range rdf:resource="xs:string"/>
 </owl:DatatypeProperty>

 <owl:DatatypeProperty rdf:ID="Engine">
  <rdfs:domain rdf:resource="#Car"/>
  <rdfs:range rdf:resource="xs:string"/>
 </owl:DatatypeProperty>
```

The second possible code fragment is the following:

```
<owl:Class rdf:ID="Car">
 <owl:unionOf rdf:parseType="Collection">
  <owl:Class rdf:about="#Wheel"/>
  <owl:Class rdf:about="#Engine"/>
 </owl:unionOf>
</owl:Class>
```

There are several ways to overcome the obstacle of rule transformation multivariance to ontology code fragments. The first way, let us name it an administrative way, is to pinpoint the kind of rule and its code fragment, which can be generated from this rule. The disadvantage of this way is that each new rule can discover imprecision and lack of optimality in previous generated code fragments. Thus, the second way, let us name it an evolutionary way, develops the first way and envisages to correct early generated code after each new rule has been transformed to code fragment. The first way will be considered first, and therefore it is necessary to determine rules and the OWL code fragments that are generated from these rules.

In the case when there is rule, where rule premis is the determining for rule conclusion, it is possible to generate a class (rule conclusion) and the properties of this class (rule premis). For example, there is the following rule:

**IF** Wings **and** Engine **THEN** Plane

It is possible to get the following OWL code fragment from this rule:

```
<owl:Class rdf:ID="#Plane"/>
 <owl:DatatypeProperty rdf:ID="Wings">
  <rdfs:domain rdf:resource="#Plane"/>
  <rdfs:range rdf:resource="xs:string"/>
 </owl:DatatypeProperty>

 <owl:DatatypeProperty rdf:ID="Engine">
  <rdfs:domain rdf:resource="#Plane"/>
  <rdfs:range rdf:resource="xs:string"/>
 </owl:DatatypeProperty>
```

In the case when there is rule, where premis consists of equivalent objects, but conclusion of the rule consists of some objects, which belongs to one of equivalent object, it is possible to generate two classes with the same properties and with one "equivalent" relation between these classes. For example, there is such rule:

**IF** (Bike **equivalent** Bicycle) **and** (Wheel, Rudder ∈ Bike)
      **THEN**   (Wheel, Rudder ∈ Bicycle)

It is possible to generate OWL code fragment from this rule, but this code fragment will depend on whether the objects of rule premis and conclusion are in being constructed OWL ontology or not. If objects of rule premis and conclusion are not in being constructed ontology, or if this rule is the first rule, which is transforming to ontology, then the code fragment is the following:

```
<owl:Class rdf:ID="Bicycle">
 <owl:DatatypeProperty rdf:ID="Wheel">
  <rdfs:domain rdf:resource="#Bicycle"/>
  <rdfs:range rdf:resource="xs:string"/>
 </owl:DatatypeProperty>
 <owl:DatatypeProperty rdf:ID="Rudder">
  <rdfs:domain rdf:resource="#Bicycle"/>
  <rdfs:range rdf:resource="xs:string"/>
 </owl:DatatypeProperty>
</owl:Class>
<owl:Class rdf:ID="Bike">
 <owl:DatatypeProperty rdf:ID="Wheel">
  <rdfs:domain rdf:resource="#Bike"/>
  <rdfs:range rdf:resource="xs:string"/>
 </owl:DatatypeProperty>
 <owl:DatatypeProperty rdf:ID="Rudder">
  <rdfs:domain rdf:resource="#Bike"/>
  <rdfs:range rdf:resource="xs:string"/>
 </owl:DatatypeProperty>
 <owl:equivalentClass>
  <owl:Class rdf:ID="Bicycle"/>
 </owl:equivalentClass>
</owl:Class>
```

It is obvious that if ontology already contains some code fragments, which are generated from previous rules and which are identical to showed OWL code fragment, the showed code fragment must not have repeating OWL code fragments.

In the case when there is rule, whose premis consists of one object, but conclusion consists of two different objects, it is possible to generate two classes and relation between these classes. For example, there is the following rule:

**IF** Driver **THEN**   hasVechicle Car

It is possible to generate such OWL code fragment from this rule:

```
<owl:Class rdf:ID="#Driver"/>
<owl:Class rdf:ID="#Car"/>
<owl:ObjectProperty rdf:ID="hasVechicle">
 <rdfs:domain rdf:resource="#Driver"/>
 <rdfs:range rdf:resource="#Car"/>
</owl:ObjectProperty>
```

In the case when there is rule, where premis consists of one object, but conclusion of the rule consists of two objects, one of which is "part_of", it is possible to generate two classes and the class of premis is a subclass of the class from the rule conclusion. For example, there is the following rule:

**IF** Wings **THEN part_of** Plane

It is possible to generate the following OWL code fragment from this rule:

```
<owl:Class rdf:ID="Wings">
 <rdfs:subClassOf>
  <owl:Class rdf:ID="Plane"/>
 </rdfs:subClassOf>
</owl:Class>
```

In the case when there is rule, where rule premis consists of one object, but rule conclusion consists of two object, one of which is „not", it is possible to generate two classes with one „not" relation between these classes. For example, there is the following rule:

**IF** Car **THEN not** Plane

It is possible to generate the following OWL code fragment from this rule:

```
<owl:Class rdf:ID="Plane"/>
<owl:Class rdf:ID="Car">
 <owl:complementOf rdf:resource="#Plane"/>
</owl:Class>
```

After all available rules are transformed to OWL ontology code fragments, using so called administrative way to overcome the obstacle of rule transformation multivariance, it is possible to tap so called evolutionary way of rule transformation multivariance overcoming in order to precise generated OWL ontology code. Here it is important to differ the rule transformation multivariance, based on the Web Ontology language redundancy of means of expression and the rule transformation multivariance, based on the accuracy of the rule information mapping. If multivariance by reason of OWL redundancy is corrected by dint of elimination of this redundancy (see administrative way), then multivariance by reason of the rule information mapping accuracy is corrected by means of precision of generated OWL code. This raises the question: is the precision of generated ontology really necessary? The answer is not so monosemantic, as it seems. Of course, at best, generated OWL ontology has to be as precise, as it is possible, but in real life it is necessary to compare quality and costs, because huge costs may produce a modest improvement in the quality. This exactly happens in the process of OWL ontology generation from rules. For example, one of the problem may occur during OWL code generation from the following rule type:

**IF** John **THEN** Man

This rule can be transformed to OWL code differently. From the human point of view it is clear that "man" is a class of many men and John is the name of some concrete man. It would be logical to represent this difference in OWL code that is to represent "Man" as a class, but "John" as an instance of the "Man" class. Any human understands this difference due to its experience, which is fixed in human's mind. The problem is that computer does not make difference among "John" and "Man" that is why it generates two classes instead of a class and its instance as it should be. A class and its instance would be an ideal variant for generation in OWL code in this case, however the development of the ability to differ classes from their instances in the computer program is rather difficult process, because it implies to know all possible names of humans. Is the precision of what is a class and what is an instance so necessary? It seems that the correct answer is no. It is quite possible to represent all objects (classes and instances) as classes. It is much more important to know what objects are and what links are among them. A class is the same instance and the difference between a class and an instance in its dimension. In other words a class is a set of instances. In turn, displaying such ontology, which has classes only and without instances, for human can be successfully implemented if to take into account and to exploit human's experience and his mind, which in the fragment of OWL ontology, showed here as a phrase:

John is a Man,

will recognize what is a class and what is an instance. In such a case accuracy of generated ontology is not essential. There is one more problem, which can arise in the process of ontology generation from rules. This problem can be arised by reason of different possible rule notations that can differ from the rule notation, used in this paper. For example, the rule, represented in this paper, uses one notation:

**IF** Wheel **and** Engine **THEN** Car,

but it can be represented, using another notation. For example, another notation can be the following:

**IF** $x \subset$ Wheel **and** Engine **THEN** $x =$ Car,

where x is some object.
It is quite possible that there are plenty of other rule notations, but it is not so important, because there is one way to cope with the problem. In these cases the problem can be solved by means of one rule notation leading to known rule notation. This is a technical task and that is why it is not discussed in this paper.

V. CONCLUSION

This paper introduces data, information, and knowledge concepts. It quotes and crarifies existing definitions of these concepts from different authors, and also gathers criticism of them. Further, an attempt of new conception of data, information and knowledge development, based on the explored disadvantages of these concepts, is made. After that, the task of OWL ontology generation from rules, which is inverse to the task of rule generation from OWL ontology, is presented. This is made by means of showing different rule types and OWL code fragments that can be generated from these rules. This work was not performed exhaustively in the sense that there were other rule types, which could be transformed to OWL code fragments, however this work was not done. This was so based on the fact that there was no urgent need to implement ontology generation algorithm from rules for the SWES (Semantic Web Expert System). SWES is expected to be an expert system, which will look for ontologies from the Web, generate rules from these ontologies, supplement its knowledge base by the ruls and reason, based on these rules and the user's request [16]. Thus, the task of rule generation from OWL ontology is much more important for the SWES, and that is why it has already been developed. In turn, the development of the task of ontology generation from rules is for the future. Fundamental ability of ontology generation from rules can be useful in order to transform "electronic" knowledge that is the knowledge, collected in computers namely in different computer expert systems, to map ontologies of different domains.

Regarding the development of the SWES, it has to be said that the task of rule generation from Web pages that is from the plain text is more important, because such an ability will give the SWES the possibility to collect knowledge from the Web in incomparably larger quantities than if the knowledge is collected from ontologies. This is so, because nowadays ontologies are not so widely circulated in the Web. This fact forces us to develop such an algorithm, which allows leveling the disadvantage.